\documentclass[10pt,twocolumn,letterpaper]{article}

\usepackage{cvpr}              

\usepackage[dvipsnames]{xcolor}
\usepackage{comment}

\newcommand{\OURS}{DIRECT-3D\xspace}

\definecolor{cvprblue}{rgb}{0.21,0.49,0.74}
\usepackage[pagebackref,breaklinks,colorlinks,citecolor=cvprblue]{hyperref}
\usepackage[accsupp]{axessibility} 

\usepackage{multirow}

\def\paperID{3125} 
\def\confName{CVPR}
\def\confYear{2024}

\title{\OURS: Learning Direct Text-to-3D Generation on Massive Noisy 3D Data}

\author{
Qihao Liu$^1$ \qquad Yi Zhang$^1$ \qquad Song Bai$^2$ \qquad Adam Kortylewski$^{3,4}$ \qquad Alan Yuille$^1$ \\
{\normalsize $^1$Johns Hopkins University \quad $^2$ ByteDance \quad  $^3$Max Planck Institute for Informatics \quad $^4$University of Freiburg}\\
\\[-1em]
\href{https://direct-3d.github.io/}{https://direct-3d.github.io/}
}

\begin{document}

\twocolumn[{%
\renewcommand\twocolumn[1][]{#1}%
\maketitle
\begin{center}
    \centering
    \vspace{-1.5em}
    \captionsetup{type=figure}
    \includegraphics[width=\linewidth]{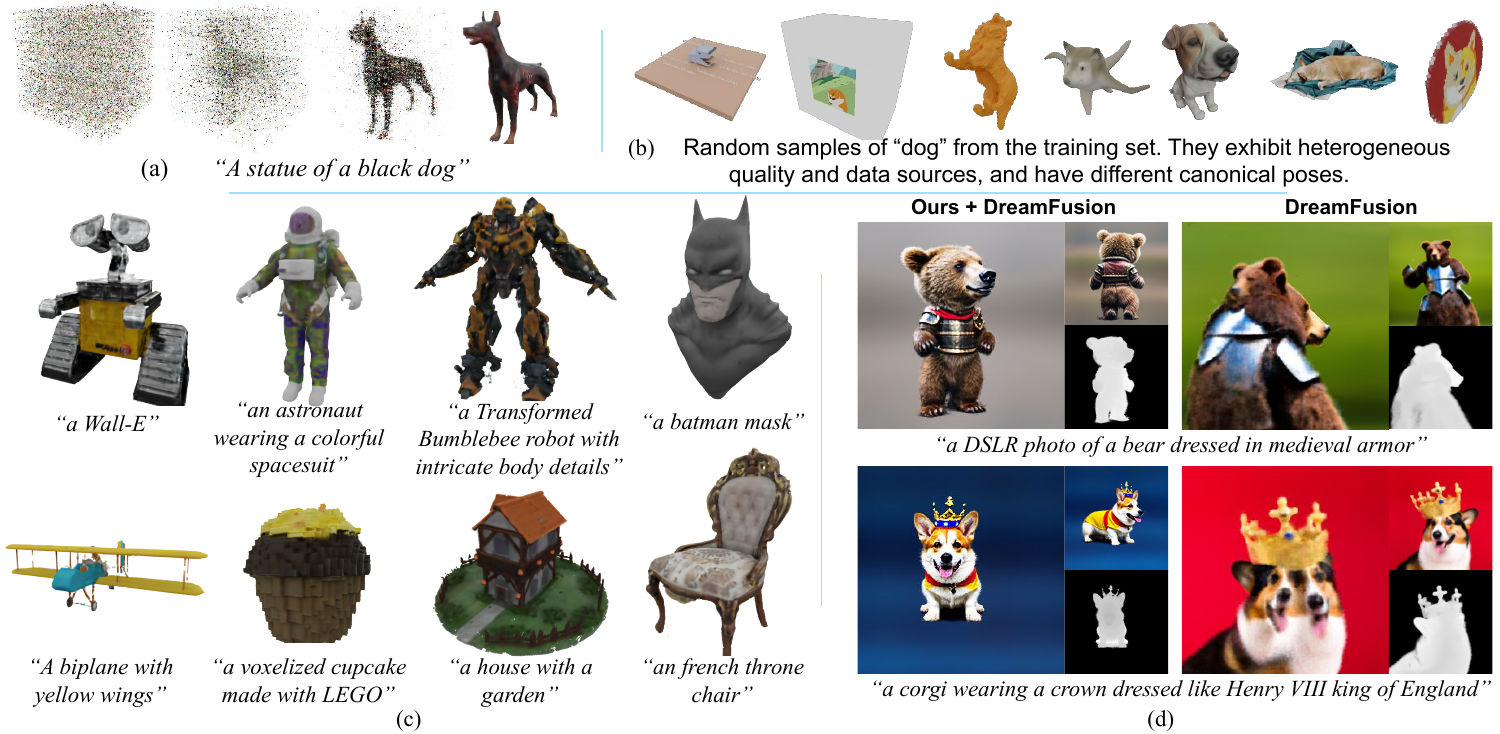}
    \captionof{figure}{ 
    Different from optimization-based 2D-lifting methods such as DreamFusion~\cite{dreamfusion}, \OURS directly generates 3D contents in a single forward pass (a).
    To mitigate the lack of high-quality 3D data, \OURS enables efficient end-to-end training of \textit{3D generative models} on massive noisy and unaligned \textit{`in-the-wild'} 3D assets (b). 
    Once trained, \OURS can generate high-quality 3D objects with accurate geometric details and various textures in 12 seconds on a single V100, driven by text prompts (c). 
    \OURS can also be used as effective 3D geometry prior that significantly alleviates the Janus problem in 2D-lifting methods (d).
    }
    \label{fig:teaser}
\end{center}%
}]
\begin{abstract}
    We present \OURS, a diffusion-based 3D generative model for creating high-quality 3D assets (represented by Neural Radiance Fields) from text prompts.
    Unlike recent 3D generative models that rely on clean and well-aligned 3D data, limiting them to single or few-class generation, our model is directly trained on extensive noisy and unaligned `in-the-wild' 3D assets, mitigating the key challenge (i.e., data scarcity) in large-scale 3D generation.
    In particular, \OURS is a tri-plane diffusion model that integrates two innovations: 1) A novel learning framework where noisy data are filtered and aligned automatically during the training process.
    Specifically, after an initial warm-up phase using a small set of clean data, an iterative optimization is introduced in the diffusion process to explicitly estimate the 3D pose of objects and select beneficial data based on conditional density.
    2) An efficient 3D representation that is achieved by disentangling object geometry and color features with two separate conditional diffusion models that are optimized hierarchically. 
    Given a prompt input, our model generates high-quality, high-resolution, realistic, and complex 3D objects with accurate geometric details in seconds.
    We achieve state-of-the-art performance in both single-class generation and text-to-3D generation.
    We also demonstrate that \OURS can serve as a useful 3D geometric prior of objects, for example to alleviate the well-known Janus problem in 2D-lifting methods such as DreamFusion.
    The code and models are available for research purposes at: \href{https://github.com/qihao067/direct3d}{https://github.com/qihao067/direct3d}.
\end{abstract}
    
\section{Introduction}
\label{sec:intro}
Diffusion models~\cite{ddpm,song2020score} have achieved significant success in 2D image synthesis~\cite{sdm,imagen,balaji2022ediffi}, owing to the large amount of image-text pairs and scaleable framework.
However, applying diffusion models to the 3D domain is challenging, mostly due to the lack of 3D data: 
Current 3D datasets are orders of magnitude smaller than their 2D counterparts, and also exhibit significant disparities in quality and complexity.
Specifically, the most widely-used dataset (\ie ShapeNet~\cite{shapenet}) comprises only 51K 3D models and focuses on individual objects.
Larger datasets like Objaverse~\cite{objaverse} and Objaverse-XL~\cite{objaversexl}, despite containing over 10M objects from Sketchfab, are noisy in quality and lack alignment (\ie, objects in varying poses).
As clean and well-aligned data continue to be very important for current methods~\cite{shue20233d,diffrf,ssdnerf}, people have to rely on high-quality yet small datasets like ShapeNet for training, and no previous 3D generative model can be directly trained on larger `in-the-wild' 3D data such as Objaverse.
As a result, these models are constrained to single-class generation, and can only generate objects with limited diversity and complexity, such as cars and tables.
In addition, the lack of efficient network design poses additional challenges, as there is no consensus on 3D data representation or network architecture that can efficiently handle high-dimensional 3D data.


To circumvent the shortage of 3D data and efficient architectures, one line of work~\cite{dreamfusion,lin2023magic3d} leverages image priors from 2D diffusion models to optimize a Neural Radiance Field (NeRF)~\cite{nerf}.
However, they are time-consuming and fragile, and often lack of geometric consistency, leading to the Janus problem (\eg, multiple faces on an animal).
Recently, one important step was made by Shap-E~\cite{shape} that directly models the distribution of large-scale 3D objects for implicit 3D representation generation.
However, they do not address the aforementioned strict requirement for training data.
Instead, they rely on vast amounts of proprietary data, which is time-consuming and costly to obtain, and they still need to invest considerable efforts to further enhance data quality~\cite{pointe}.
In addition, Shap-E necessitates multi-stage training with a complex recipe, requiring point clouds and RGBA images with per-pixel 3D coordinates as input.


In this work, we present \OURS, a \textbf{D}iffusion model with \textbf{I}te\textbf{R}ativ\textbf{E} optimization for \textbf{C}onditional \textbf{T}ext-to-\textbf{3D} generation (Fig.~\ref{fig:teaser}).
It enables direct training on massive noisy and unaligned \textit{`in-the-wild'} 3D data in an end-to-end manner, with multi-view images as supervision.
Given a text prompt, it generates a variety of high-quality 3D objects (NeRFs) with precise geometric details and diverse textures within seconds.
Our model consists of a 2D diffusion module to generate tri-plane features~\cite{eg3d} and a NeRF decoder to extract NeRF parameters from the generated tri-plane. 
Tri-plane features facilitate an efficient 3D representation in well-established 2D networks, and NeRF offers an effective and compact way to model intricate details of 3D objects.
To tackle the aforementioned challenges, we made the following important technical innovations:


\textbf{Firstly}, we incorporate an iterative optimization process into the diffusion step to explicitly estimate the pose and quality of the 3D data based on the conditional density of the diffusion model, enabling automatic cleaning and alignment of the data during training.
It considerably reduces the need for high-quality and precisely aligned 3D data and opens up a novel method  to efficiently train 3D generative models on large amounts of \textit{`in-the-wild'} 3D assets.
\textbf{Secondly}, we disentangle 3D geometry and 2D color of the object, modeling them hierarchically with two separate diffusion models. 
The geometry tri-plane is generated first, and the color is generated conditioned on geometry and the text prompt.
This disentanglement enhances the efficiency and capabilities for modeling 3D data.
It also allows for more flexible usage of our model. 
For example, our geometry diffusion module can be seamlessly integrated in existing Score Distillation Sampling~\cite{dreamfusion} based approaches, and provide additional 3D geometry priors, which significantly improve the geometry consistency while preserving the high-fidelity texture from the 2D image diffusion models.
\textbf{Finally}, we propose an automated method to generate multiple descriptive prompts for each object, spanning from coarse to fine-grained levels, which enhances the alignment between prompt features and the generated 3D objects.


We evaluate \OURS on both single-class generation and text-to-3D generation. 
For single-class generation, our method outperforms all previous methods on all tested categories by a large margin when trained on exactly the same data (\eg, from 14.27 to 7.26 in FID), proving our effectiveness in modeling 3D data. 
For text-to-3D generation, we achieve superior performances compared to previous work (Shap-E~\cite{shape}), excelling in quality, detail, complexity, and realism.
User studies show that $73.9\%$ of raters prefer our approach over Shap-E.
In addition, when used as geometry prior, our method significantly improves the 3D consistency of previous 2D-lifting models (\eg DreamFusion~\cite{dreamfusion}), and raises the generation success rate from $12\%$ to $84\%$.

In summary, we make the following contributions:
\begin{itemize}
\item We propose \OURS, which enables end-to-end training of 3D generative models on extensive noisy and unaligned `in-the-wild' 3D data. It achieves state-of-the-art performance on both single-class and large-scale text-guided 3D generation.

\item Given text prompts, \OURS is able to generate high-quality, high-resolution, realistic, and complex 3D objects (NeRFs) with precise geometric details in seconds.

\item \OURS provides important and easy-to-use 3D geometry prior of arbitrary objects, complementing 2D priors provided by image diffusion models.

\end{itemize}

\begin{figure*}
    \centering
    \includegraphics[width=\linewidth]{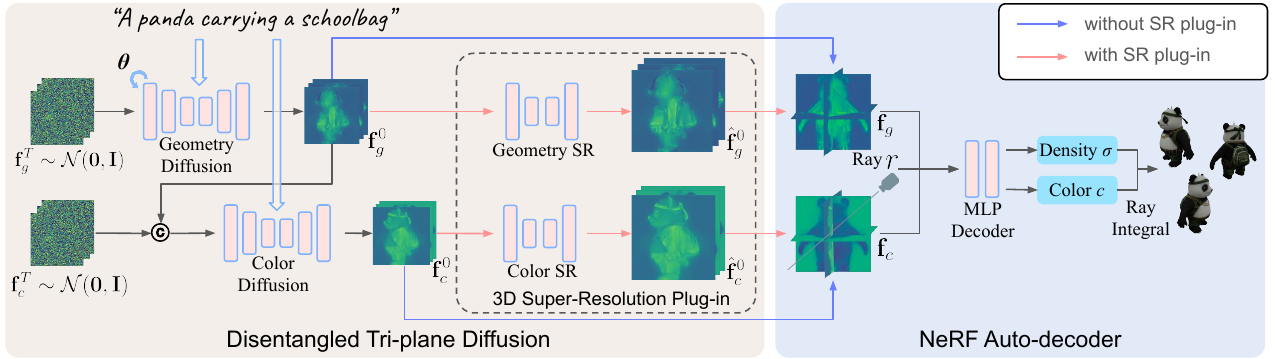}
    \caption{\textbf{Method overview.} 
    Given a prompt, we generate a NeRF with two modules: The disentangled tri-plane diffusion module uses 2 (or 4 if the super-resolution plug-in is used) diffusion models to generate geometry ($\mathbf{f}_g$) and color ($\mathbf{f}_c$) tri-plane separately. 
    Then both tri-planes are reshaped and fed into a NeRF auto-decoder to get the final outputs. 
    During training, an iterative optimization process is introduced in the geometry diffusion to explicitly model the pose $\theta$ of objects and select beneficial ones, enabling efficient training on noisy `in-the-wild' data. 
    The whole model is end-to-end trainable (with or without SR plug-in), with only multi-view 2D images as supervision. }
    \vspace{-1.0em}
    \label{fig:pipeline}
\end{figure*}

\section{Related Work}
\label{sec:relatedWork}

\noindent\textbf{Direct 3D generation.}
Early work relies on either GAN~\cite{gan} or VAE~\cite{vae} to model the distribution of 3D objects, represented by voxel grids~\cite{brock2016generative,wu2016learning,gadelha20173d}, point clouds~\cite{achlioptas2018learning,yang2019pointflow,zhou20213d,mo2019structurenet}, or implicit representations~\cite{park2019deepsdf,sitzmann2020implicit,chen2019learning}. 
Recently, diffusion models~\cite{ddpm,song2020score} have been utilized to create objects with appearance~\cite{gupta20233dgen,diffrf,anciukevivcius2023renderdiffusion,karnewar2023holodiffusion,karnewar2023holofusion,ssdnerf} or pure geometric shapes~\cite{luo2021diffusion,zhou20213d,hui2022neural,zheng2023locally,zeng2022lion,cheng2023sdfusion,erkocc2023hyperdiffusion,li2023diffusion,shue20233d,zhang20233dshape2vecset}.
However, these methods are constrained by their reliance on clean and well-aligned 3D datasets such as ShapeNet~\cite{shapenet}. 
Hence, they can only focus on a single category or a few categories.

Recently, Cao~\etal~\cite{cao2023large} train a class-conditional 3D diffusion model on OmniObject3D~\cite{omni3d}, which contains 216 object categories, enabling large-vocabulary 3D generation. 
However, their need for well-aligned 3D data limits their training set to just 5.9K objects, averaging only 27 objects per category, which severely restricts the quality and diversity.
To enable large-scale 3D generation, Point-E~\cite{pointe} and Shap-E~\cite{shape} train text-conditional diffusion models on massive proprietary data. 
However, acquiring such data is costly and time-consuming, and large efforts are still required to further enhance the data quality~\cite{pointe}.
In contrast, we directly tackle this key constraint on training data by enabling direct training on extensive `in-the-wild' 3D data, which is cost-effective and easy to scale up.

\noindent\textbf{Text-to-3D generation with 2D diffusion.} 
To circumvent the constraints imposed by limited 3D data and enable large-scale generation, another line of work~\cite{dreamfusion,wang2023score,lin2023magic3d,metzer2023latent,wang2023prolificdreamer,chen2023fantasia3d,tsalicoglou2023textmesh,huang2023dreamtime} leverages pre-trained 2D image diffusion priors for 3D generation. 
However, they are known for suffering from the Janus problem, in which radially asymmetric objects exhibit unintended symmetries, due to the lack of 3D consistency in 2D diffusion models.
MV-Dream~\cite{shi2023mvdream} mitigates this issue by fine-tuning a pre-trained image diffusion model to produce multi-view images, highlighting the importance of 3D knowledge.
In contrast, we directly generate objects in 3D space with accurate geometry information.
Moreover, our method provides accurate 3d geometry priors to these 2D-based methods, complementing the 2D priors from image diffusion models, and hence effectively alleviating the Janus problem.
In addition, these methods require tens of minutes to hours for optimizing a single object, whereas our method generates NeRFs in seconds.

\section{Method}
\label{sec:method}

Our model consists of a tri-plane diffusion module to generate tri-planes of a 3D object, and a NeRF auto-decoder~\cite{park2019deepsdf} to decode the tri-planes into final radiance field. 
In Sec.~\ref{secMethod:arch}, we introduce our architecture design. 
Sec.~\ref{secMethod:em} describes how we can train our model on noisy and unaligned 3D data. 
In Sec.~\ref{secMethod:sr}, we introduce the 3D super-resolution plug-in for high-resolution generation. 
Sec.~\ref{secMethod:prompt} describes an automated way to generate descriptive captions in different granularities.
Training and implementation details are available in the Supp.
An overall illustration is provided in Fig.~\ref{fig:pipeline}.

\subsection{Tri-plane Diffusion for NeRF Generation}
\label{secMethod:arch}

\noindent\textbf{NeRF generation from disentangled tri-plane representation.}
Given a set of 2D multi-view images of a subject, one can learn its 3D representation with a NeRF, which models the subject using volume density $\sigma\in \mathbb{R}_+$ and RGB color $c\in \mathbb{R}_+^3$.
For a more efficient representation, we follow previous work~\cite{rodin,ssdnerf} that uses the tri-plane representation to model the NeRFs. 
Specifically, it factorizes a 3D volume into three axis-aligned orthogonal 2D feature planes $\mathbf{f}_{xy},\mathbf{f}_{xz},\mathbf{f}_{yz}\in\mathbb{R}^{N\times N \times C}$. 
Then, one can query the feature $\mathbf{f}$ of any 3D point $p\in\mathbb{R}^3$ by projecting it onto each of the three planes and aggregating the retrieved features.


However, we find it necessary to disentangle the geometry and color features into two separate tri-planes, denoted by $\mathbf{f}_g$ and $\mathbf{f}_c$ respectively, which improves model capability and provides important geometry prior (see Sec.~\ref{sec:abl:disen}).
Then, with the tri-planes $\mathbf{f}_g$ and $\mathbf{f}_c$, and a set of rays $\{r_i\}$, we can get the integral radiance $y$ of this subject with an auto-decoder: $y_{i} = \mathcal{R}(\mathcal{D}_\omega(\mathbf{f}_g,\mathbf{f}_c,r_i))$, where $\mathcal{D}_\omega$ is a multi-layer perceptron decoder with parameters $\omega$, $\mathcal{R}$ denotes volume rendering~\cite{max1995optical}, and $i$ is the ray index. 
Our decoder processes the tri-planes $\mathbf{f}_g$ and $\mathbf{f}_c$ separately to generate density and color, thereby ensuring that $\mathbf{f}_g$ only encapsulates the geometry information and $\mathbf{f}_c$ only contains the corresponding color features (see Supp. for details).
Given the ground-truth pixel RGB $\hat{y}$, the tri-planes $\mathbf{f}_g$, $\mathbf{f}_c$ and parameters $\omega$ can be optimized by minimizing the rendering loss:
\begin{align}
    \mathcal{L}_{rad}(\mathbf{f}_g, \mathbf{f}_c,\omega) = \sum_i||\hat{y}_i - \mathcal{R}(\mathcal{D}_\omega(\mathbf{f}_g,\mathbf{f}_c,r_i))||^2_2
    \label{eqn:lrad}
\end{align}

\noindent\textbf{Disentangled tri-plane generation.}
For conditional generation of tri-plane $\mathbf{f}_{(\cdot)}$ from prompt $p$, we adopt a 2D latent diffusion model~\cite{ddpm,sdm}. In our framework, the diffusion model denoises tri-plane features $\mathbf{f}_g, \mathbf{f}_c \in\mathbb{R}^{N\times N \times 3C}$ that stack the channels of all three axes into a single image.

Given an input tri-plane $\mathbf{f}_g^0$ (or $\mathbf{f}_c^0$), the diffusion model progressively adds noise to it and produces a noisy output $\mathbf{f}_g^t:=\alpha^t\mathbf{f}_g^0+\sigma^t\epsilon$ at timestep $t$, where $\epsilon\sim\mathcal{N}(\mathbf{0},\mathbf{I})$ is the added Gaussian noise, $\alpha^t$ and $\sigma^t$ are noise schedule functions.
During each training step, we first train a geometry denoising network $\epsilon_\phi(\mathbf{f}_g^t,t,\tau(p))$ via
\begin{align}
    \mathcal{L}_{geo}(\phi) = \mathbb{E}_{\mathbf{f}_g^0,\epsilon,p,t}[||\epsilon - \epsilon_\phi(\mathbf{f}_g^t,t,\tau(p))||^2_2]
    \label{eqn:lgeo}
\end{align}
where $\tau$ denotes a pre-trained CLIP text encoder~\cite{clip}.
Then, a color denoising network $\epsilon_\psi(\mathbf{f}_c^t,t,\tau(p),\mathbf{f}_g)$ conditioned on both prompt $p$ and geometry $\mathbf{f}_g$ is optimized by
\begin{align}
    \mathcal{L}_{col}(\psi) = \mathbb{E}_{\mathbf{f}_c^0,\epsilon,p,\mathbf{f}_g,t}[||\epsilon - \epsilon_\psi(\mathbf{f}_c^t,t,\tau(p),\mathbf{f}_g)||^2_2]
\end{align}
Prompt condition is added by a cross-attention mechanism~\cite{sdm} with classifier-free guidance~\cite{ho2022classifier}, and geometry condition for color diffusion is added via concatenation.

During inference, the geometry tri-plane $\mathbf{f}_g^0$ is sampled starting from the Gaussian noise $\mathbf{f}_g^T\sim\mathcal{N}(\mathbf{0},\mathbf{I})$ conditioned on prompt $p$, then the color tri-plane $\mathbf{f}_c^0$ is sampled similarly but conditioned on prompt $p$ and geometry $\mathbf{f}_g^0$.

\subsection{Training with Noisy and Unaligned Data}
\label{secMethod:em}

Beyond our disentangled architecture and the introduced training objective, large-scale text-to-3D synthesis requires a substantial amount of 3D data for training.
Recent efforts~\cite{objaverse, objaversexl} have gathered over 10M `in-the-wild' 3D objects from Sketchfab.
However, these datasets are difficult to use due to the heterogeneous quality and data sources, and the lack of alignment, leading to poor performance or even non-convergence during training (see Sec.~\ref{sec:abl:em}).
Manual cleaning and alignment of 10M data is time-consuming and impractical to scale up.
To this end, we introduce an iterative optimization process within the diffusion training step to autonomously identify noisy 3D data and automatically align clean data samples during training.  

To achieve this goal, for each object, we explicitly model its 3D rotation angle as $\theta = \{\theta_\mu,\theta_\sigma\}$, where $\theta_\mu,\theta_\sigma\in\mathbb{R}^3$ denote the estimated mean and variance of its 3D rotation angle. 
Once estimated, the rotation angle can be sampled from $\mathcal{N}(\theta_\mu,\theta_\sigma)$. 
Note that the geometry tri-plane $\mathbf{f}_g$ is now conditioned on the rotation $\theta$, so Eqn.~\ref{eqn:lgeo} becomes
\begin{align}
    \mathcal{L}_{geo}(\phi,\theta) = \mathbb{E}_{\mathbf{f}_g^0;\theta,\epsilon,p,t}[||\epsilon - \epsilon_\phi(\mathbf{f}_g^t;\theta,t,\tau(p))||^2_2]
    \label{eqn:lgeoEM}
\end{align}
Then, we can estimate the rotation parameter $\theta$ by also minimizing the diffusion loss $\mathcal{L}_{geo}(\phi,\theta)$. 
However, directly minimizing it w.r.t $\theta$ is challenging, since our model only uses multi-view images as supervision, and the tri-plane reconstruction already requires hundreds of optimization iterations per object (although effectively). 
Note that we do not need an accurate estimate of $\theta$; instead, a rough pose with good axis disentanglement in tri-plane suffices (see Fig.~\ref{fig:abl_em_triplane}).

To perform this estimation, we consider $\theta$ as a hidden variable and propose an iterative optimization process. 
We first initialize the model with a very short warm-up phase on a small aligned dataset (details in Supp.). 
Subsequently, during each training iteration on the entire noisy dataset, we sample $m$ different $\theta$ following $\mathcal{N}(\theta_\mu,\theta_\sigma)$ and estimate the corresponding tri-planes $\mathbf{f}_g^0$. 
Then with a frozen geometry diffusion model, we compute the loss in Eqn.~\ref{eqn:lgeoEM} with fixed parameter $\phi$ at a fixed time step $t$, which gives us a loss distribution w.r.t the sampled rotations $\theta$. 
After that, we can update the rotation parameter by $\theta_\mu \leftarrow (1-\lambda_\mu)\theta_\mu +\lambda_\mu\theta_{min}$ and $\theta_\sigma \leftarrow \lambda_\sigma|\theta_\mu-\theta_{min}|$, where $\theta_{min}$ is the sampled rotation with the smallest loss and $\lambda_{(\cdot)}$ are momentum parameters.
Finally, given threshold $T$, we can use $\theta_{min}$ to update the geometry denoising network $\epsilon_\phi$ if $\mathcal{L}_{geo}(\phi,\theta_{min}) \leq T$.

We initialize $\theta_\mu$ and $\theta_\sigma$ with all elements equal to $0$ and $\pi$, respectively. 
Then we set $m=\texttt{ceil}(36/\pi\cdot\theta_\sigma)$, which is updated every iteration. 
In practice, it converges fast, often requiring just 5-10 iterations.
We filter out the objects that do not converge after 10 iterations. 
This step does not require back-propagation through the diffusion model when optimizing $\theta$, which also speeds up the process.

\subsection{3D Super Resolution}
\label{secMethod:sr}
Directly training a high-resolution diffusion model is slow and inefficient.
Therefore, we train our base module at a resolution of $128^2$, and rely upon a 3D Super-Resolution (SR) plug-in with the tri-plane diffusion structure to increase the resolution from $128^2$ to $512^2$. 
Given a low-resolution tri-plane $\mathbf{f}_{(\cdot)}$, we first apply a roll-out operation~\cite{rodin} that concatenates the tri-plane features horizontally, followed by a bilinear interpolation to get an intermediate tri-plane $\mathbf{f}'_{(\cdot)}$ at a resolution of $512^2$.
Then, a parameterized diffusion model is used to directly predict the high-resolution tri-plane $\mathbf{\hat{f}}_{(\cdot)}$.
Alongside the L2 loss on tri-plane, we apply an entropy loss to the generated NeRF to encourage full transparent or opaque points, ensuring a smoother SR generation.
It's worth noting that \textbf{our model can directly generate high-quality objects without the SR plug-in.}
In fact, except for results in Fig.~\ref{fig:teaser} (c), all experiments/results in this paper are conducted \textbf{without} the SR module to ensure fair comparisons with baselines, as they are all evaluated at $128^2$.
More details are provided in the Supp.

\subsection{Coarse to Fine-gained Caption Generation}
\label{secMethod:prompt}
Text prompts play a crucial role in large-scale generation, but datasets like Objaverse only contain paired metadata that do not serve as informative captions.
To solve this problem, Cap3D~\cite{cap3d} proposed to use LLM to consolidate captions generated from multiple views of a 3D object. 
We follow their pipeline to generate captions for all training examples. 
However, we found that these captions may be overly detailed and contain irrelevant objects, making it difficult to train a model from scratch.
In addition, considering the limited availability of 3D data, we find that caption enrichment with different granularities is an effective and cost-efficient manner to `scale up' the training set.

To generate more accurate captions with multiple granularities, we begin by rendering 8 images at $512^2$ from different camera angles for each object. 
Next, a pretrained DeiT~\cite{deit} on ImageNet-1K~\cite{deng2009imagenet} is used to classify the object in each image and output object proposals based on the top-5 confidence scores. 
After that, we use BLIP2~\cite{li2023blip} and LLaVA~\cite{liu2023visual} for captioning through a two-stage question-answering process. 
In the first stage, they are tasked to identify the object in the image. 
Then we compare the identified object with the object proposals using the CLIP similarity, and eliminate irrelevant objects. 
In the second stage, for each image, the top-ranked matched answer is passed to the vision-language models for (1) assigning a title to this object, and providing descriptions of the object's (2) color and texture, and (3) structure and geometry.
5 answers are generated for each question.  
Then we adopt the caption selection and consolidation from Cap3D~\cite{cap3d} to get the final captions. 
We retain four captions per object, which correspond to (1) the object category, (2) the generated title, and the descriptions focusing on (3) texture and (4) geometry. 
Finally, we use the category and title information to further eliminate the irrelevant objects in descriptions (3) and (4).
These captions are selected randomly during training.

\section{Experiments}
\label{sec:experiments}

In this section, we first evaluate the performance of our method on single-class generation (Sec.~\ref{sec:exp:sin}) and large-scale text-to-3D generation (Sec.~\ref{sec:exp:text}). 
Then, we show that our method can function as a critical object-level 3D geometry prior, significantly improving previous optimization-based text-to-3D models (Sec.~\ref{sec:exp:prior}).
Finally, we prove the effectiveness of our main ingredients in ablation (Sec.~\ref{sec:exp:abl}). 
Additional experimental results are provided in the Supp.

\noindent\textbf{Datasets.}
We warm up our model on OmniObject3D~\cite{omni3d} and a split of ShapeNet~\cite{shapenet}, which contain 6342 objects spanning 216 categories.
Then we train our full model on Objaverse~\cite{objaverse} that contains 800K+ objects.\footnote{We did not use Objaverse-XL~\cite{objaversexl} since the data were not public available when this project was conducted.} 
\textit{For single-class generation}, we strictly follow the previous methods~\cite{dupont2022data,diffrf,ssdnerf} and conduct experiment on ShapeNet SRN Cars~\cite{shapenet}, Amazon Berkeley Objects (ABO) Tables~\cite{collins2022abo}, and PhotoShape (PS) Chairs~\cite{park2018photoshape}. 
For Chairs, we generate images following the render pipeline in DiffRF~\cite{diffrf}.
For Cars and Tables, we directly use the rendered images in SSDNeRF~\cite{ssdnerf} for both training and testing.

\begin{figure*}
    \centering
    \includegraphics[width=\linewidth]{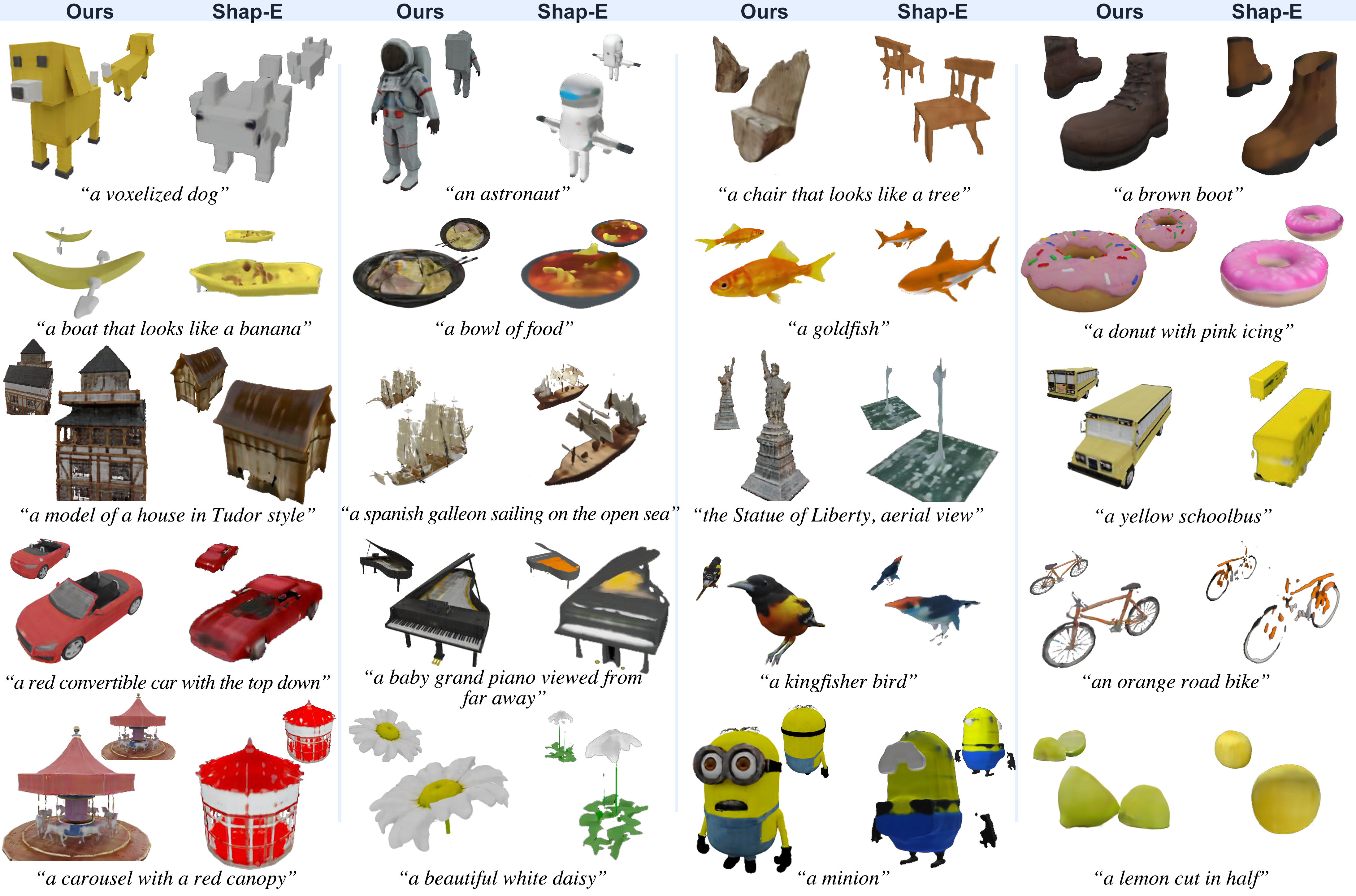}
    \caption{\textbf{Qualitative comparison with Shap-E~\cite{shape}.} 
    We use the same text prompt as in Shap-E (top 2 rows) and DreamFusion (middle 2 rows), we also compare the performance on complex objects (last row). 
    For Shap-E, we use the official code and model.
    For our method, we generate objects in $128^3$ without the super-resolution plug-in.
    All images of both methods are rendered at $256^2$.
    Our \OURS generates 3D objects with enhanced quality in both geometry and texture.    
    We also generate more various and complex objects.
    }
    \vspace{-1.0em}
    \label{fig:shap-e}
\end{figure*}

\subsection{Single-class 3D Generation}
\label{sec:exp:sin}
\begin{table}[]
\setlength{\lightrulewidth}{0.01em}
\setlength{\cmidrulewidth}{0.01em}
\setlength\tabcolsep{3pt}
\resizebox{1\columnwidth}{!}{
\begin{tabular}{lcccccc}
\toprule
\multirow{2}{*}{Method} & \multicolumn{2}{c}{Car} & \multicolumn{2}{c}{Chair} & \multicolumn{2}{c}{Table} \\ \cmidrule(lr){2-3}  \cmidrule(lr){4-5}  \cmidrule(lr){6-7} 
                        & FID ($\downarrow$)& KID ($\downarrow$)& FID ($\downarrow$)& KID ($\downarrow$) & FID ($\downarrow$) & KID ($\downarrow$)          \\ \midrule
$\pi$-GAN~\cite{chan2021pi}       & 36.7        & -         & 52.71       & 13.64       & 41.67       & 13.82       \\
EG3D~\cite{eg3d}           & 10.46       & 4.90      & 16.54       & 8.41        & 31.18       & 11.67       \\
DiffRF~\cite{diffrf}        & -           & -         & 15.95       & 7.94        & 27.06       & 10.3        \\
SSDNeRF~\cite{ssdnerf}     & 11.08       & 3.47      & -           & -           & 14.27       & 4.08        \\ \midrule
Ours                  & \textbf{6.90}    & \textbf{1.84}  & \textbf{7.01}    & \textbf{2.12}    & \textbf{7.26}   & \textbf{1.89}       \\ \bottomrule
\end{tabular}}
\caption{\textbf{Single-class generation on SRN Cars, PS Chairs, and ABO Tables.} 
Baseline results are reported by DiffRF and SSDNeRF. 
We train our model from scratch using exactly the same rendered images as the baselines.
KID is multiplied by $10^3$. }
\label{tab:single_class}
\end{table}

We compare against four leading methods: $\pi$-GAN~\cite{chan2021pi}, EG3D~\cite{eg3d}, DiffRF~\cite{diffrf}, SSDNeRF~\cite{ssdnerf}.
Following the latest SOTA method (SSDNeRF), we evaluate the generation quality using the Fr\'{e}chet Inception Distance (FID)~\cite{heusel2017gans} and Kernel Inception Distance (KID)~\cite{binkowski2018demystifying}. 
All metrics are evaluated at a resolution of $128^2$.
Results are reported in Tab.~\ref{tab:single_class}.
We reduce our model size to $135$M parameters for a fair comparison with SSDNeRF ($122$M).
We also remove the prompt condition and train a separate model on each category following the baselines.
\textbf{Even when trained from scratch on the same data with a similar model size, our approach significantly outperforms all previous methods.}
It underscores the high quality of our generated objects and the effectiveness of our method in modeling 3D data.
Qualitative comparisons are provided in the Supp.

\subsection{Direct Text-to-3D Generation}
\label{sec:exp:text}
\begin{table}[]
\setlength{\lightrulewidth}{0.01em}
\setlength{\cmidrulewidth}{0.01em}
\resizebox{1\columnwidth}{!}{
\begin{tabular}{lccc}
\toprule
 & More realistic & More detailed & Overall preference \\ \midrule
Shap-E~\cite{shape}    & 28.4\%    &  22.9\%    &    26.1\%    \\
Ours               &   71.6\%      &  77.1\%     &   73.9\%     \\ \bottomrule
\end{tabular}}
\caption{\textbf{User preference studies.} We conduct user studies on 475 prompts, including all prompts from Shap-E and 162 prompts from DreamFusion. 73.9\% of users prefer ours over Shape-E.}
\label{tab:user}
\end{table}

We compare our method with the current SOTA method Shap-E~\cite{shape}.
For a fair and comprehensive comparison, we evaluate both methods on 475 prompts, including all prompts in the official paper and website of Shap-E and 162 prompts from DreamFusion gallery.~\footnote{https://dreamfusion3d.github.io/gallery.html}
Qualitative results are provided in Fig.~\ref{fig:shap-e}. 
\textbf{Our model is able to generate more various and complex objects with much higher quality in both geometric details and textures. }
More results can be found in the Supp.

Following Magic3D~\cite{lin2023magic3d}, we also conduct user studies to evaluate different methods based on user preferences on Amazon MTurk.
For each generated object, we render a video recording its rotation along the z-axis, covering a full 360-degree view. 
Then we show users two side-by-side videos generated by two algorithms, both using the same input prompt. 
We randomly switch the order of these two videos for different prompts.
Users are instructed to evaluate which video is (1) more realistic, (2) more detailed, and (3) which one they prefer overall.
Each prompt is evaluated by 3 different users, yielding a total of 1425 comparison results.
As shown in Tab.~\ref{tab:user}, \textbf{we generate more realistic and detailed objects, leading to higher user preference.}

\subsection{Improving 2D-lifting Methods with 3D Prior}
\label{sec:exp:prior}

\begin{figure*}
    \centering
    \includegraphics[width=\textwidth,height=10cm]{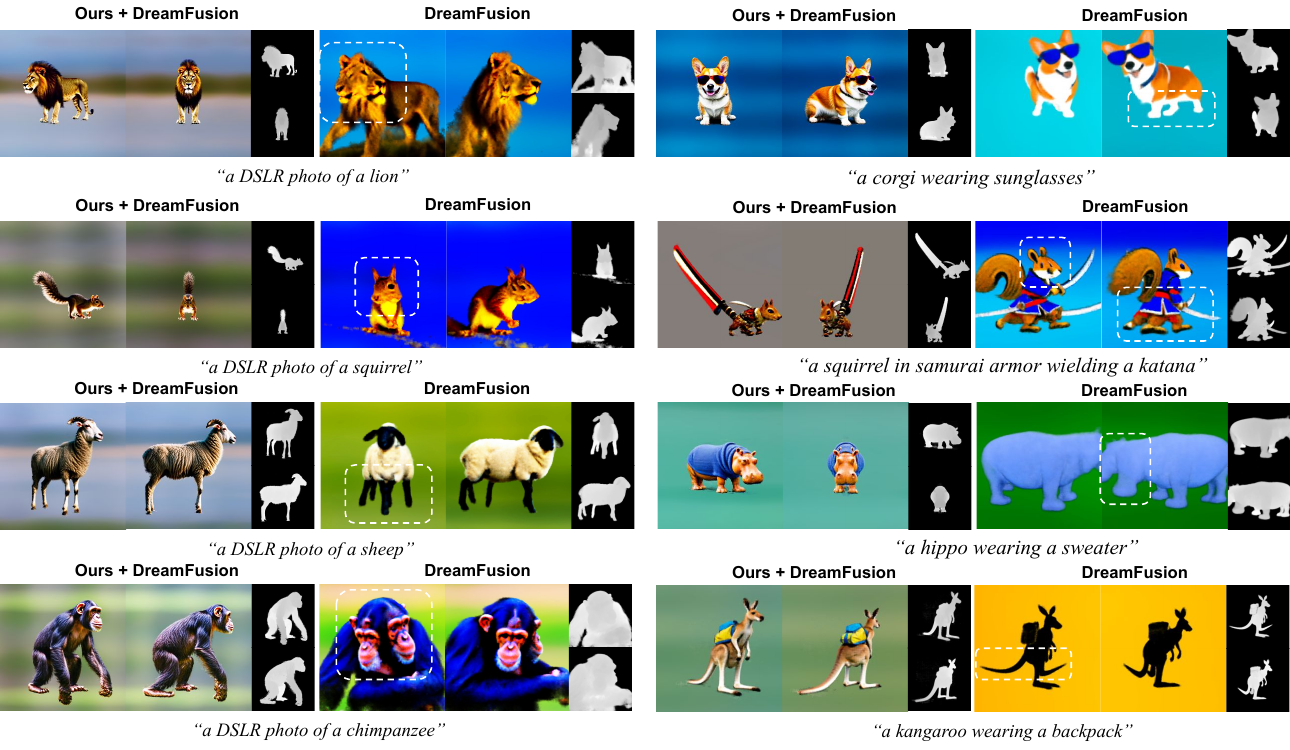}
    \caption{\textbf{\OURS provides a useful 3D prior for 2D-lifting methods~\cite{dreamfusion}.} 
    Our 3D prior alleviates issues such as multiple faces and missing/extra limbs, while also improving texture quality.
    Please check the video results in Supp. for a better comparison.
    }
    \label{fig:dreamfusion_cmp}
    
\end{figure*}

Recent 2D-lifting text-to-3D methods~\cite{dreamfusion,lin2023magic3d} have demonstrated impressive visual quality and compositionality using pretrained 2D text-to-image diffusion models as image prior. 
However, they suffer from the multi-face (Janus) problem. 
Here we show that plugging \OURS into the 2D-lifting framework as a 3D prior greatly alleviates the Janus problem and improves the geometry consistency.

We use an open-source implementation of DreamFusion~\cite{dreamfusion} using StableDiffusion v2.1~\cite{Rombach_2022_CVPR} (DreamFusion-SD) or DeepFloyd~\cite{deepfloyd} (DreamFusion-IF) as the 2D image prior. 
Our 3D prior is implemented as a Score Distillation Sampling (SDS)~\cite{dreamfusion} loss added to the original text-to-3D loss. 
As the Janus problem only happens on radially asymmetric objects like animals, we concentrate our quantitative experiments on animals. 
We conducted 50 trials using the prompt `A DSLR photo of a \texttt{[animal]}', with \texttt{[animal]} randomly sampled from a list of 14 animal types.
The prompt for \OURS is set to `A \texttt{[animal]}'. 
Only generations with both correct geometry and texture are counted as success. 
The detailed criterion is described in the Supp. 
As shown in Tab.~\ref{tab:imp_dre}, \textbf{adding \OURS as 3D prior greatly improves the success rate of text-to-3D generation, alleviating the multi-face problem.}

\begin{table}[]
\setlength{\lightrulewidth}{0.01em}
\setlength{\cmidrulewidth}{0.01em}
\resizebox{1\columnwidth}{!}{
\begin{tabular}{lccc}
\toprule
 & Succ. Rate & Geo. Consist. & Tex. Consist. \\ \midrule
DreamFusion-SD~\cite{dreamfusion}        &     12\%            &        16\%     &      30\%            \\
DreamFusion-IF~\cite{dreamfusion}        &      10\%       &      10\%       &     72\%                 \\
DreamFusion-SD $+$ Ours               &       \textbf{84\%}      &     \textbf{84\%}      &    \textbf{98\%}          \\ \bottomrule
\end{tabular}}
\caption{\textbf{Improving 2D-lifting text-to-3D generation.} \OURS provides a useful 3D geometry prior, enhancing the geometry consistency and increasing the generation success rate.}
\vspace{-1.0em}
\label{tab:imp_dre}
\end{table}

We also show qualitative comparisons in Fig~\ref{fig:dreamfusion_cmp}. Our method provides important geometry prior that greatly improves the generation success rate and the geometry consistency of the baseline method. In addition, we find that with better geometry information, the texture consistency and quality are also improved.

\subsection{Ablation Studies}
\label{sec:exp:abl}

\subsubsection{Ablation of Automatic Alignment and Cleaning}
\label{sec:abl:em}
\begin{table}[]
\setlength{\lightrulewidth}{0.01em}
\setlength{\cmidrulewidth}{0.01em}
\setlength\tabcolsep{3pt}
\resizebox{1\columnwidth}{!}{
\begin{tabular}{lcccccc}
\toprule
 & \multicolumn{2}{c}{Car (R)} & \multicolumn{2}{c}{Chair (R)} & \multicolumn{2}{c}{Car + Chair + Table (R)} \\ \cmidrule(lr){2-3}  \cmidrule(lr){4-5}  \cmidrule(lr){6-7} 
                        & FID ($\downarrow$)& KID ($\downarrow$)& FID ($\downarrow$)& KID ($\downarrow$) & FID ($\downarrow$) & KID ($\downarrow$)          \\ \midrule
w/o AAC                    &  46.77       &  34.35     & 45.57         & 27.17        &   39.06     &  27.24       \\
w/ AAC                    &  8.69     &  2.82     & 10.53        &  5.35      &  13.62      & 5.03   \\ \bottomrule
\end{tabular}}
\caption{\textbf{Automatic Alignment and Cleaning (AAC) improves performance on unaligned data.} To simulate unaligned data, all objects are rotated by a random degree, with a maximum of $360^{\circ}$ along z-axis and $\pm 30^{\circ}$ along x/y axes (denoted as R).
C+C+T means a same model is trained on all 3 datasets for multi-class generation, with `A 3D mesh of a \texttt{[Class]}' as prompt condition.}
\vspace{-1.0em}
\label{tab:abl_em}
\end{table}

\begin{figure}
    \centering
    \includegraphics[width=0.9\linewidth]{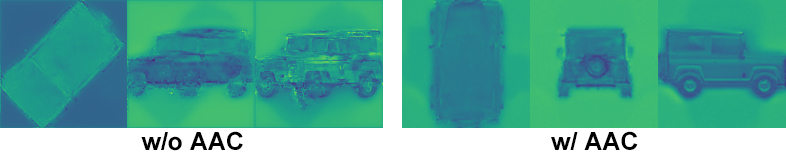}
    \caption{\textbf{Tri-plane feature learned with/without Automatic Alignment and Cleaning (AAC) on Objaverse.} 
    It roughly aligns the objects to get clear tri-plane features.
    Unaligned objects can be captured by tri-plane representation, but the inadequate axis disentanglement makes it challenging for the diffusion model to learn.
    }
    \vspace{-0.5em}
    \label{fig:abl_em_triplane}
\end{figure}

\begin{figure}
    \centering
    \includegraphics[width=0.98\linewidth]{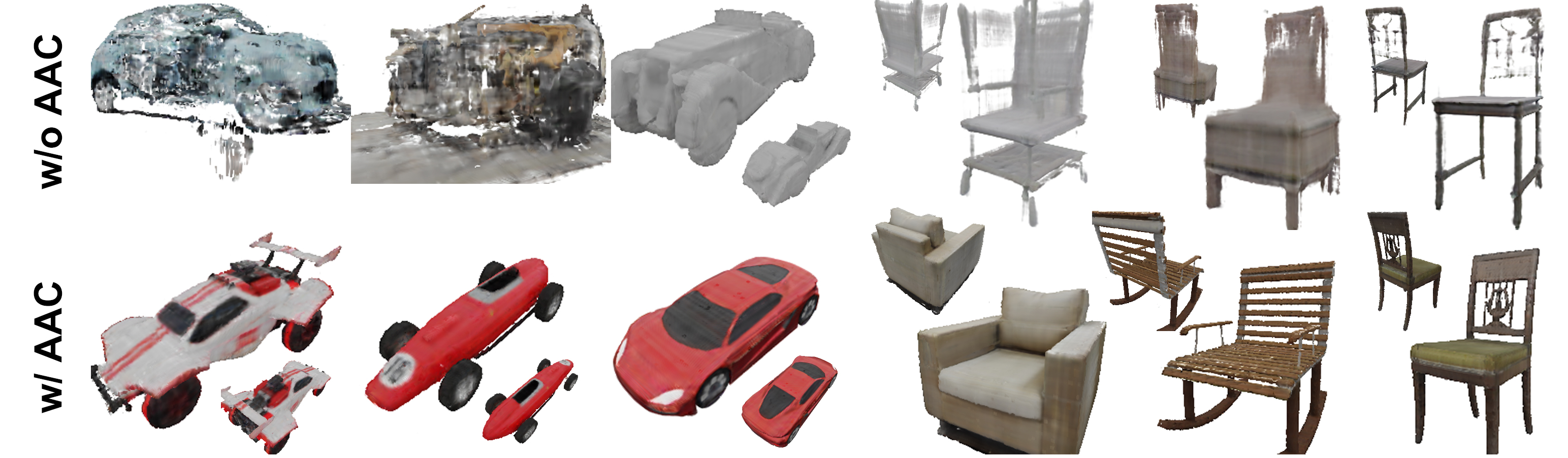}
    \caption{\textbf{Model learned with/without AAC on Objaverse.}
    AAC enables direct and more efficient training on noisy, unaligned data.
    }
    \vspace{-1.0em}
    \label{fig:abl_em_mesh}
\end{figure}

We show the effectiveness of the Automatic Alignment and Cleaning (AAC) in Tab.~\ref{tab:abl_em}, Fig.~\ref{fig:abl_em_triplane}, and Fig.~\ref{fig:abl_em_triplane}.
For \textit{quantitative evaluation}, we randomly rotated the aligned objects in SRN Cars, ABO Tables, and PS Chairs, and evaluate the models on their test set. 
Results are provided in Tab.~\ref{tab:abl_em}.
For \textit{visualization}, we select cars and chairs from the Objaverse dataset based on their assigned category title, and directly train our model on them.
We visualize the learned tri-plane and the generated NeRFs in Fig.~\ref{fig:abl_em_triplane} and Fig~\ref{fig:abl_em_mesh}.
AAC learns reasonable alignments of 3D objects while effectively filtering out toxic data.
It enables direct and more efficient training on noisy and unaligned `in-the-wild' data.

\vspace{-0.5em}
\subsubsection{Ablation of Disentanglement}
\label{sec:abl:disen}
\begin{table}[]
\setlength{\lightrulewidth}{0.01em}
\setlength{\cmidrulewidth}{0.01em}
\setlength\tabcolsep{3pt}
\resizebox{1\columnwidth}{!}{
\begin{tabular}{lccccccc}
\toprule
 & \multicolumn{2}{c}{Car} & \multicolumn{2}{c}{Table} & \multicolumn{2}{c}{Car + Chair + Table} \\ \cmidrule(lr){2-3}  \cmidrule(lr){4-5}  \cmidrule(lr){6-7} 
                        & FID ($\downarrow$)& KID ($\downarrow$)& FID ($\downarrow$)& KID ($\downarrow$) & FID ($\downarrow$) & KID ($\downarrow$)          \\ \midrule
Not Disentangled        & 9.98       & 2.96        & 12.86         & 3.87            &   17.74    &  8.15      \\
Disentangled            & 6.90       & 1.84          & 7.26        & 1.89            &   10.06      &  3.44        \\ \bottomrule
\end{tabular}}
\vspace{-0.5em}
\caption{\textbf{Improvement of Disentanglement.}}
\vspace{-0.5em}
\label{tab:abl_disentangle}
\end{table}
Tab.~\ref{tab:abl_disentangle} highlights the enhancements achieved through disentanglement.
For models without disentanglement, we double the number of layers to maintain similar model parameters.
Disentanglement greatly improves model capabilities, establishing the foundation for large-scale generation.

\begin{figure}
    \centering
    \includegraphics[width=0.98\linewidth]{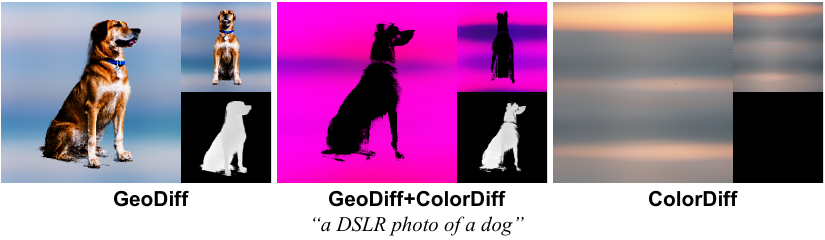}
    \vspace{-1.0em}
    \caption{\textbf{Disentangling geometry and color provides a proper 3D geometrical prior, while improving the high-fidelity texture from 2D image diffusion models.}}
    \label{fig:dreamfusion_disentangle}
    \vspace{-1.0em}
\end{figure}

More importantly, it provides pure geometry priors for various tasks.
Considering 2D-lifting text-to-3D generation,
Fig.~\ref{fig:dreamfusion_disentangle} shows that when geometry and color are \textit{not} disentangled, using our model as a geometry prior also affects the texture (\ie, harms the image feature prior learned from 2D diffusion models).
However, with disentanglement, we are able to provide critical geometry priors while preserving the high-fidelity texture from 2D image diffusion models. 
In addition, with better geometry consistency, the textures learned from 2D diffusion models are also improved.


\subsubsection{Ablation of Prompt Enrichment}

\begin{figure}
    \centering
    \includegraphics[width=\linewidth]{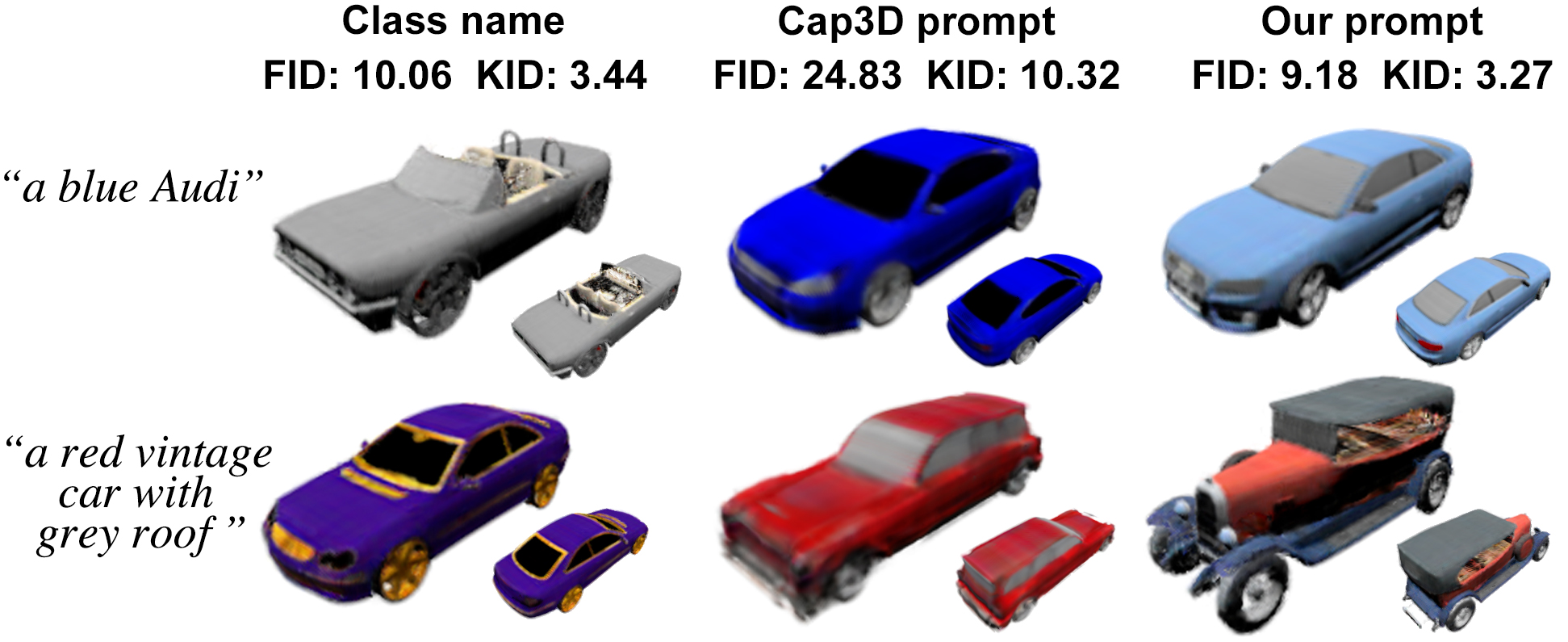}
    \caption{\textbf{Prompt Enrichment.} 
    FID and KID are computed on the entire test set.
    We provide captions with varying granularities: Coarse captions enhance object-category connections, simplifying the training, while fine-gained captions enable a better understanding of detailed features such as color and part-level information.
    }
    \vspace{-1.0em}
    \label{fig:abl_prompt}
\end{figure}

Fig.~\ref{fig:abl_prompt} compares the performance variance when the model is trained with different prompts.
`Class name' means caption with template `A 3D mesh of a \texttt{[Class]}'.

\noindent{\textbf{Class name}} gives a better performance on FID and KID scores (reported in the figure). 
It simplifies the problem into a class-conditional multi-class generation task, ensuring higher quality in the generated object.
However, training only with class names leads to a lack of basic understanding regarding detailed attributes.

\noindent{\textbf{Cap3D prompt}} contains finer details, yet can be overly intricate and occasionally contains irrelevant objects or even incorrect captions due to the failure of BLIP2 on synthetic objects.
Directly training on them is more challenging, resulting in reduced quality and lower FID/KID scores.

\noindent{\textbf{Our prompt enrichment}} provides 4 different prompts for each object under different granularities. 
It ensures high-quality generation while offering better control over details.

\section{Conclusion}
\label{sec:conclusion}

We have presented \OURS, a diffusion-based text-to-3D generation model that is directly trained on extensive noisy and unaligned ‘in-the-wild’ 3D assets.
Given text prompts, \OURS can generate high-quality 3D objects with precise geometric details in seconds.
It also provides important and easy-to-use 3D geometry priors, complementing 2D priors provided by image diffusion models.

\section*{Acknowledgement}
This work was done in part during an internship at ByteDance.
AY acknowledges support from the ONR N00014-21-1-2812 and Army Research Laboratory award W911NF2320008.
AK acknowledges support via his Emmy Noether Research Group funded by the German Science Foundation (DFG) under Grant No. 468670075.

\clearpage
\setcounter{page}{1}
\maketitlesupplementary

In this supplementary document, we provide details and extended experimental results omitted from the main paper for brevity. 
Specifically, Sec.~\ref{supp:nerf} provides details of the NeRF Auto-decoder.
Sec.~\ref{supp:3DSUP} provides details of the 3D Super-Resolution module.
Then, we cover the training details in Sec.\ref{supp::train}, including loss functions and warm-up training on clean data.
Sec.~\ref{supp:expDetails} presents experiment details and hyperparameters.
Sec.~\ref{supp:addExp} gives additional ablation studies and more qualitative results.
Finally, the limitations of our method are discussed in Sec.~\ref{supp:limit}.

In addition, we provide \textbf{video results for all visualizations in the supplementary file}.

\section{Model Details}

\subsection{NeRF Auto-decoder}
\label{supp:nerf}

\begin{figure}
    \centering
    \includegraphics[width=\columnwidth]{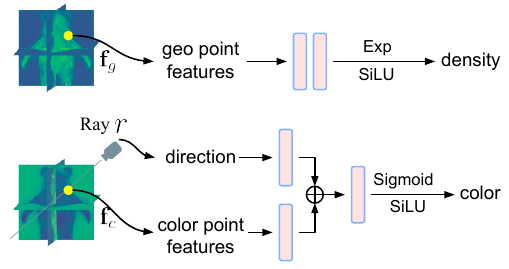}
    \caption{\textbf{Architecture of the NeRF auto-decoder.} }
    \vspace{-1.0em}
    \label{fig:nerf_Decoder}
\end{figure}

We employ a NeRF Auto-decoder to extract features from the generated tri-planes and get NeRF parameters.
This auto-decoder consists of several multi-layer perceptrons to process the tri-plane features $\mathbf{f}_g$ and $\mathbf{f}_c$ separately. 
Fig.~\ref{fig:nerf_Decoder} illustrates its architecture, which contains several fully connected layers with non-linear activation functions. 
The decoding process involves two distinct branches to handle the tri-plane features separately, ensuring that $\mathbf{f}_g$ encapsulates only the geometry information and $\mathbf{f}_c$ contains only the corresponding color features.

\subsection{3D Super Resolution}
\label{supp:3DSUP}
Similar to the structure of the base tri-plane diffusion model, the 3D super-resolution (SR) module also employs a U-Net model as its backbone.
However, we apply only one upsampling layer that directly scales the tri-plane feature from $128^2$ to $512^2$.
To enable efficient training with a larger batch size, we train the SR module separately. 
Therefore, we can directly use the saved tri-plane features during the training of the base model to train the SR module. 
Following cascaded image generation~\cite{ho2022cascaded}, we add Gaussian blurring and Gaussian noises to the intermediate tri-plane feature $\mathbf{f}'_{(\cdot)}$.

For training, alongside the L2 loss $\mathcal{L}_{geo}(\psi^{SR})$ and $\mathcal{L}_{col}(\phi^{SR})$ on tri-plane, we apply an entropy loss
\begin{align*}
    \mathcal{L}_{entropy}=\rho\cdot\text{log}_2(\rho) - (1-\rho)\cdot\text{log}_2(1-\rho)
\end{align*}
to the generated NeRF to encourage full transparent or opaque points, ensuring a smoother SR generation.
Here $\rho$ denotes the cumulative sum of density weights computed when computing NeRF parameters from tri-plane features.

\subsection{Training and Implementation Details}
\label{supp::train}

\noindent\textbf{Loss function.}
To enable larger batch size and expedite training, we first exclude the 3D super-resolution (SR) module and train the base model end-to-end at $128^2$, by minimizing the following objective:
\begin{align*}
    \mathcal{L}_{base} = \lambda_{geo}\mathcal{L}_{geo}(\phi,\theta)&+\lambda_{col}\mathcal{L}_{col}(\psi)\\
    &+\lambda_{rad}\mathcal{L}_{rad}(\mathbf{f}_g, \mathbf{f}_c,\omega)
\end{align*}
To speed up the convergence of tri-planes learned from multi-view images (\ie, $\mathcal{L}_{rad}(\mathbf{f}_g, \mathbf{f}_c,\omega)$), we adopt prior gradient caching~\cite{ssdnerf} and save the diffusion gradients $\nabla_{\mathbf{f}_g}\mathcal{L}_{geo}$ and $\nabla_{\mathbf{f}_c}\mathcal{L}_{col}$ for re-using to update the tri-plane.
It enables us to update $\mathcal{L}_{rad}(\mathbf{f}_g, \mathbf{f}_c,\omega)$ multiple times in one training iteration.

Then we freeze the base tri-plane diffusion module and only train the SR module to get high-resolution generations at $512^2$, with the following objective:
\begin{align*}
    \mathcal{L}_{SR} = & \lambda_{geo}\mathcal{L}_{geo}(\phi^{SR})+\lambda_{col}\mathcal{L}_{col}(\psi^{SR})\\
    &+\lambda_{rad}\mathcal{L}_{rad}(\mathbf{f}_g, \mathbf{f}_c,\omega)+\lambda_{entropy}\mathcal{L}_{entropy}
\end{align*}
In this step, we load, resize, and fine-tune the tri-plane features saved during the training of the base diffusion module. 
We use bilinear interpolation to scale the saved tri-planes from $128^2$ to $512^2$.

\noindent\textbf{Warm-up training.}
Training the entire system is challenging due to the intricate interdependencies between different modules. 
Specifically, optimizing diffusion model is less effective when tri-plane $\mathbf{f}_{(\cdot)}$ in Eqn. 1 is far from convergence, but learning $\mathbf{f}_g$ with rotation $\theta$ needs a reasonably functioning diffusion model.
Therefore, we warm up the model on clean and well-aligned data for the first 1/50 of the total iterations.
It also defines a universal canonical pose for all objects.
After that, we continue the training on all datasets with a learnable rotation parameter $\theta$ using the algorithm described in Sec. 3.2.

\section{Experiment Details}
\label{supp:expDetails}
\subsection{Direct Text-to-3D Generation}

We warm up our model on OmniObject3D~\cite{omni3d} and a split of ShapeNet~\cite{shapenet}, which contain 6342 objects spanning 216 categories. 
Then we train our full model on Objaverse~\cite{objaverse} that contains 800K+ objects.

\noindent\textbf{Hyperparameters.}
We first train our base model for 2M iterations with a batch size of 256. 
Then the SR module is trained for 500K iterations with a batch size of 32.
Both module are trained on 32 A100 GPUs.
We set the number of channels for tri-plane features $C=6$, and train a diffusion model with $1000$ diffusion steps with linear noise schedule to generate the tri-plane features.
During inference we sample $50$ diffusion steps.
The latent base learning rate is $1e^{-2}$ for all experiments.
The learning rates for both geometry and color diffusion models are set to $1e^{-4}$, and the learning rate for NeRF auto-decoder is set to $1e^{-3}$.
$\lambda_{geo}=\lambda_{col} = 5$, $\lambda_{rad}=20$, and $\lambda_{entropy}=0.1$.
We update the tri-plane reconstructions from multi-view images 16 times per iteration for the initial 200K training iterations, and once per iteration for the subsequent training iterations.
The latent base learning rate is reduced by a factor of $0.5$ after 500K iterations and by a factor of $0.1$ after 1M iterations.

\subsection{Single-class 3D Generation}
We reduce our model size to 135M parameters for a fair comparison with SSDNeRF~\cite{ssdnerf} (122M). 
We also remove the prompt condition and train a separate model on each category following the baselines.

\noindent\textbf{Hyperparameters.}
All models are trained for 500K iterations on 8 A100 GPUs, utilizing a batch size of 64. 
No SR plug-in is trained during these experiments.
For cars and tables, the latent base learning rate is set to $4e^{-2}$.
In the case of chairs, the latent base learning rate is set to $5e^{-3}$.
The remaining hyperparameters align with those specified in direct text-to-3D generation.

\subsection{Improving DreamFusion with 3D Prior}

In our experiments, we sample \texttt{[animal]} from 14 animal types: \textit{bear, corgi, dog, bird, cat, pig, elephant, horse, sheep, zebra, squirrel, chimpanzee, tiger, lion}. 

\noindent\textbf{Criterion for successful generation.}
We consider a text-to-3D generation successful when both the generated geometry and texture are consistent. 
Consistent geometry implies the correct number of parts is generated without missing or extra ones. 
Consistent texture implies the generated texture contains a consistent and plausible pattern that may appear on an actual animal of that type, regardless of the geometry.

\noindent\textbf{Hyperparameters.} 
For DreamFusion and \OURS, we run 10K iterations of optimization using the Adam optimizer~\cite{xie2022adan} with a learning rate of $5 \times 10^{-3}$.
Perp-Neg~\cite{armandpour2023re} is enabled for the 2D diffusion guidance with $w_{\textrm{neg}}=-4$ for all methods, which we found useful to reduce incorrect textures such as multiple head textures.
We set the weight of the 3D prior SDS loss provided by \OURS to $0.01$.
The classifier-free guidance is set to $100$ as suggested in DreamFusion~\cite{dreamfusion}. 
We use a coarse-to-fine training process for all methods, starting from a spatial resolution of $64^2$ for the first 5K iterations and increasing to $128^2$ afterward. 
The remaining hyperparameters are set to the default values. 

\section{Additional Experiments}
\label{supp:addExp}
\subsection{Ablation on the Super-resolution Module}
\begin{figure}
    \centering
    \includegraphics[width=\columnwidth]{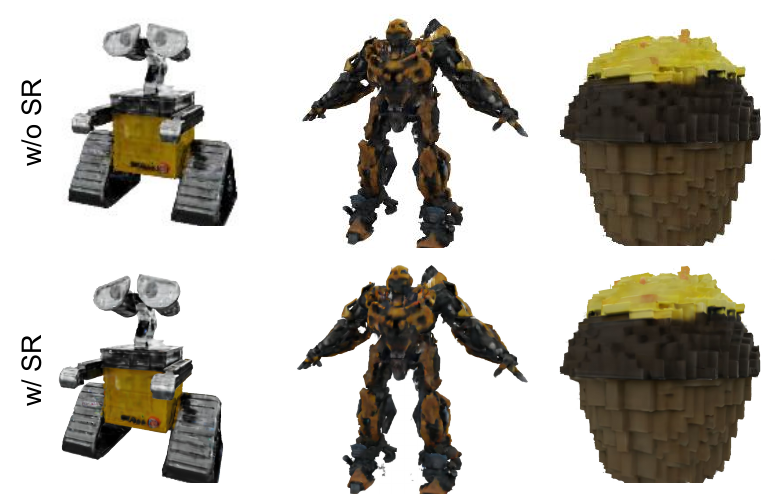}
    \caption{\textbf{Comparison of generated objects with and without the 3D super-resolution plug-in.} 
    Please zoom in for better visualization.}
    \vspace{-1.0em}
    \label{fig:supp_abl_sr}
\end{figure}

We employ an additional 3D super-resolution plug-in to enhance the resolution from $128^3$ to $256^3$. 
Fig.~\ref{fig:supp_abl_sr} compares the generated objects with and without the SR plug-in, demonstrating its effectiveness in producing high-resolution objects with reduced computational resources.
However, it's worth noting that the SR plug-in may slightly alter the generated low-resolution objects and introduce additional noise.

\subsection{Ablation on 3D Prior Loss Weight}
\begin{figure}
    \centering
    \includegraphics[width=\columnwidth]{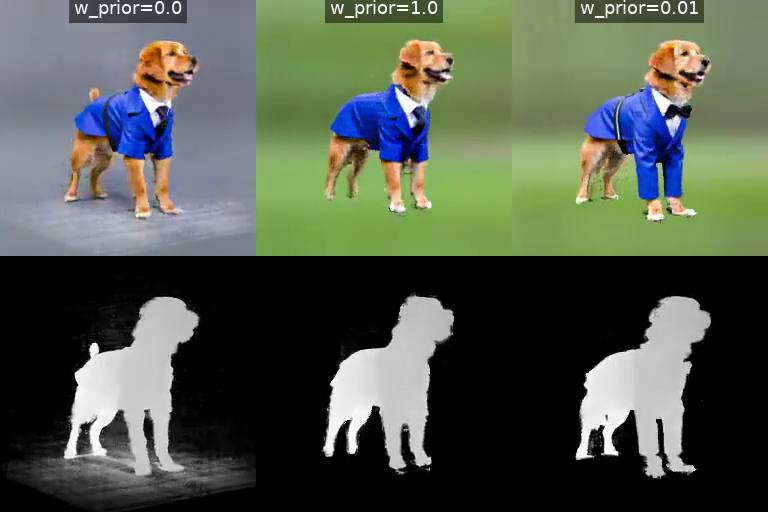}
    \caption{\textbf{Ablation of 3D prior loss weight.}}
    \label{fig:supp_abl_prior_w}
\end{figure}

We also study the impact of different 3D prior loss weights.
Ablation in Fig.~\ref{fig:supp_abl_prior_w} shows that utilizing only \OURS as initialization can alleviate the Janus problem, but also results in many artifacts, while large weights could compromise the quality of the generated geometry (\eg, missing rear feet in this case). 

\subsection{Additional Qualitative Examples}
\begin{figure*}[htbp]
    \centering
    \subfloat[EG3D]{\includegraphics[width=.9\linewidth]{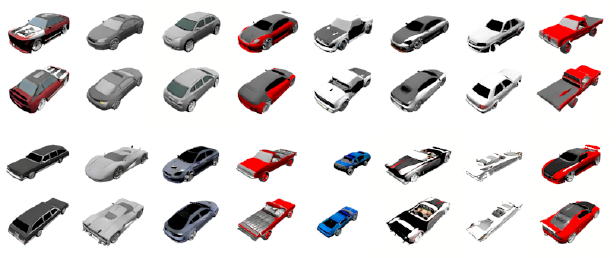}}\\
    \subfloat[SSDNeRF]{\includegraphics[width=.9\linewidth]{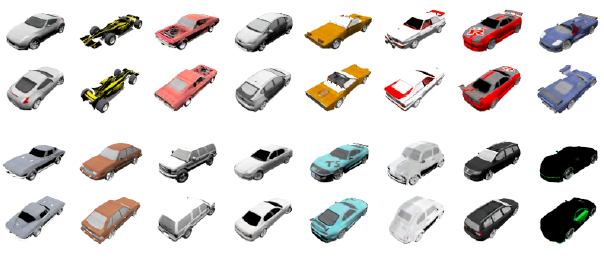}}\\
    \subfloat[Ours]{\includegraphics[width=.9\linewidth]{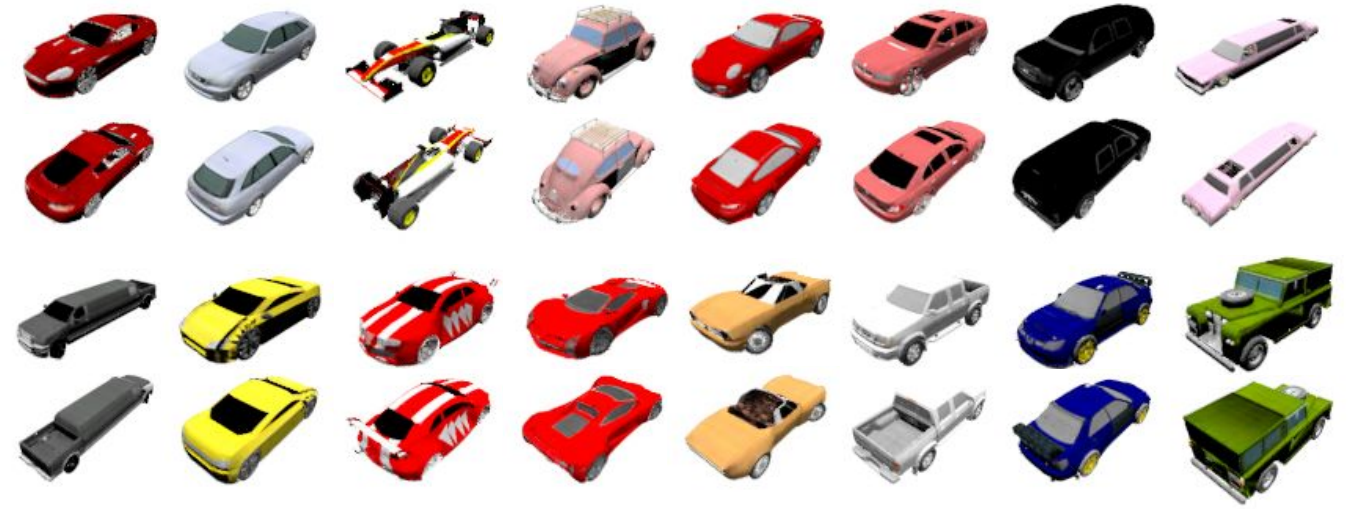}}
    \caption{\textbf{Qualitative comparison on ShapeNet SRN Cars.} 
    Baseline results come from the original paper of SSDNeRF~\cite{ssdnerf}.
    Following the baseline methods, we generate and render images at $128^2$.}
    \label{fig:supp_car}
\end{figure*}

\begin{figure*}
    \centering
    \includegraphics[width=0.98\textwidth]{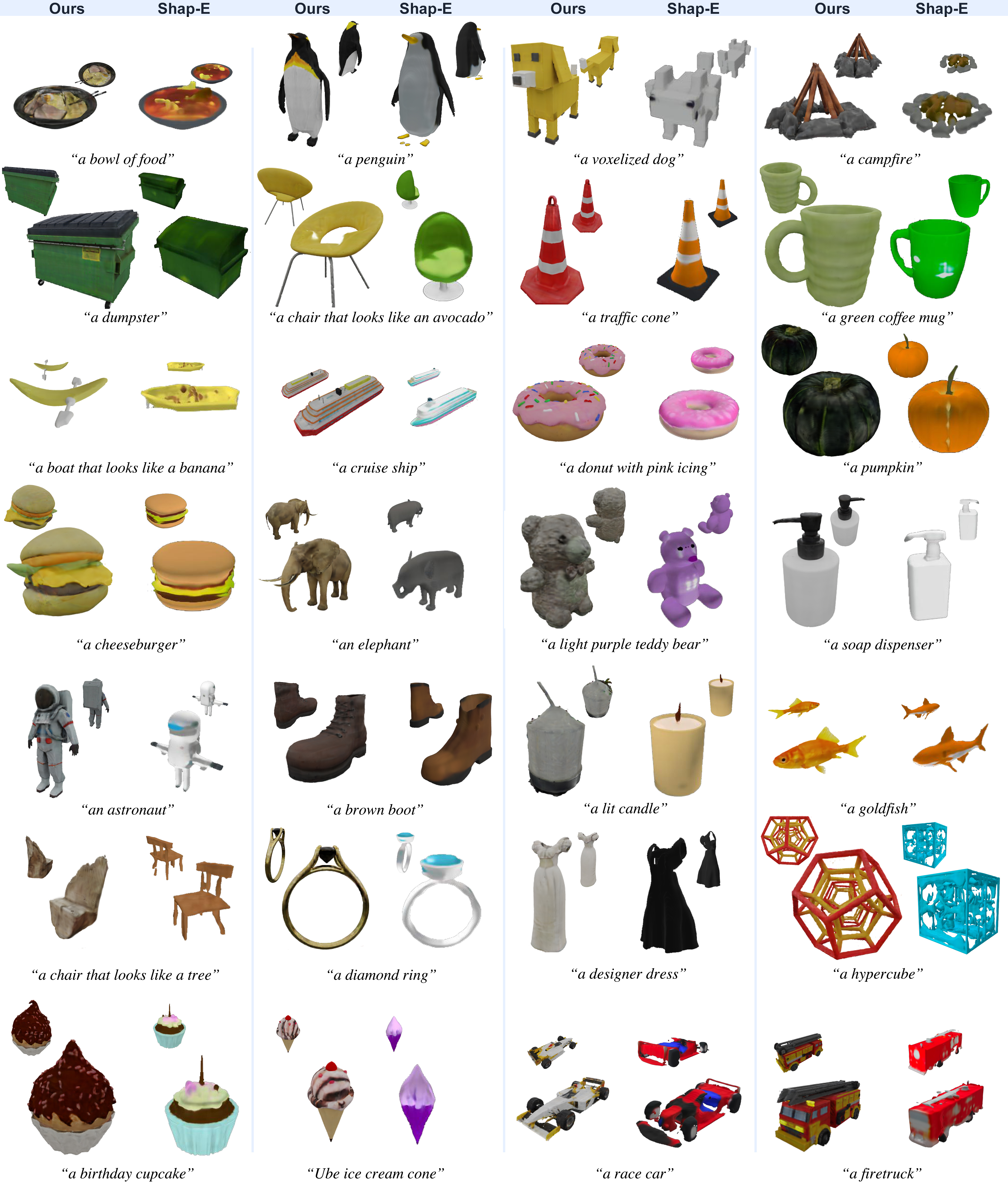}
    \caption{\textbf{Qualitative comparison with Shap-E~\cite{shape}.} 
    All text prompts are sourced from the original paper of Shap-E.
    For Shap-E, we use the official code and model with the default random seed.
    For our method, we generate objects in $128^2$ without the super-resolution plug-in. 
    All images of both methods are rendered at $256^2$. 
    Our DIRECT-3D generates 3D objects with enhanced quality in both geometry and texture. 
    }
    \label{fig:supp_shape}
\end{figure*}

\begin{figure*}
    \centering
    \includegraphics[width=0.98\textwidth]{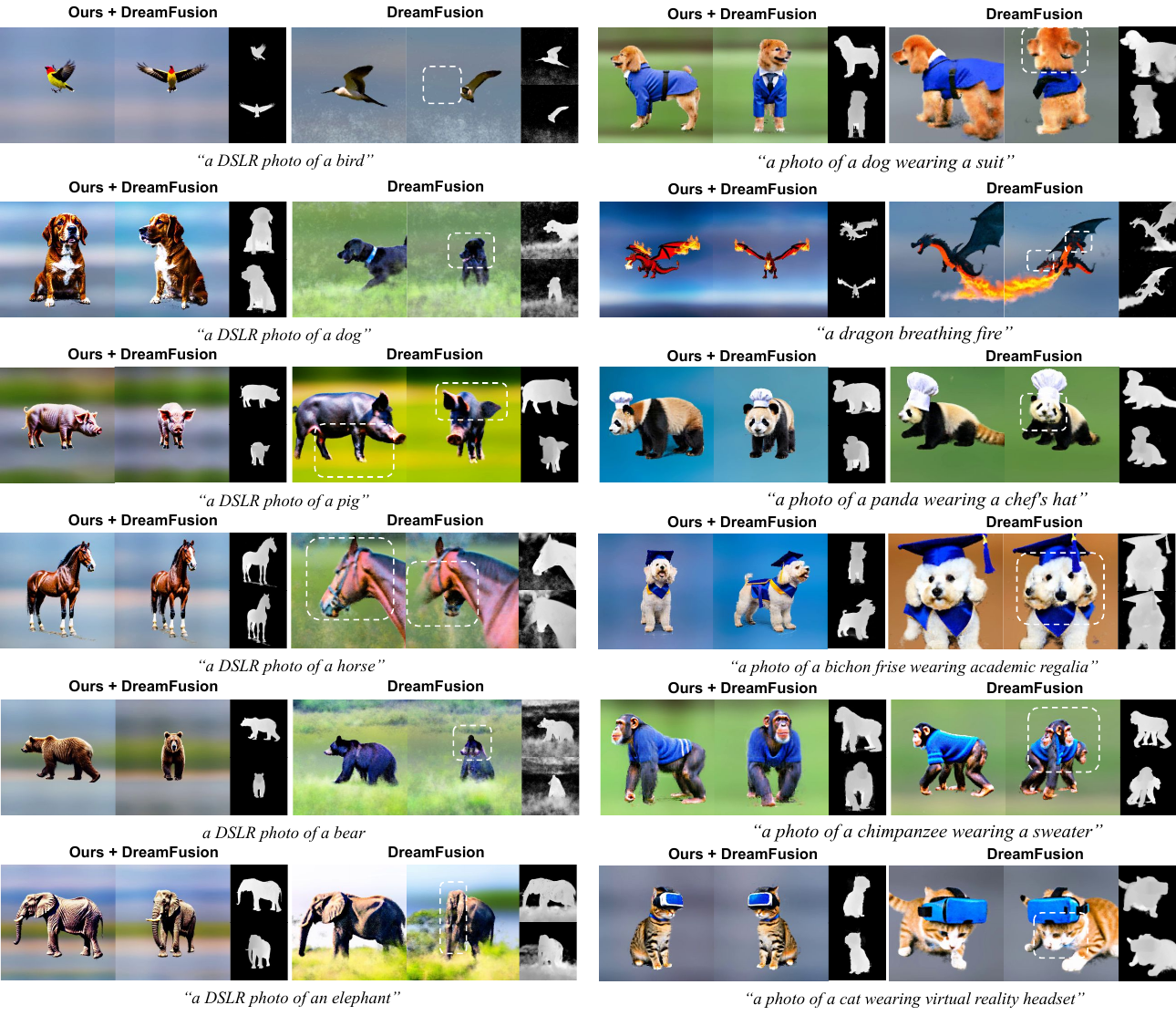}
    \caption{\textbf{More qualitative results on using \OURS as a 3D prior for 2D-lifting methods.} 
    Our 3D prior alleviates issues such as multiple faces and missing/extra limbs, while also improving texture quality. 
    Please also check the video demos for a better visualization.  
    }
    \label{fig:supp_dreamfusion}
\end{figure*}

We provide additional qualitative comparisons here. 
Specifically, Fig.~\ref{fig:supp_car} provides qualitative comparison with EG3D~\cite{eg3d} and SSDNeRF~\cite{ssdnerf} on single-class 3D generation.
Fig.~\ref{fig:supp_shape} provides additional comparisons with Shap-E~\cite{shape} on direct text-to-3D generation, using the same text prompts as in Shap-E. 
Fig.~\ref{fig:supp_dreamfusion} provides additional qualitative results on using \OURS as a 3D prior to improve 2D-lifting text-to-3D methods such as DreamFusion~\cite{dreamfusion}.

\section{Limitations}
\label{supp:limit}
While \OURS consistently produces high-quality results and surpasses previous methods in single-class 3D generation and direct text-to-3D synthesis, it does exhibit certain limitations.
First of all, despite the abundant geometry information provided by large-scale 3D datasets, a significant proportion of them lacks realistic textures. 
Additionally, the synthetic-to-real gap still persists, even for objects with nice and detailed textures. 
Therefore, training a 3D generative model, such as \OURS, solely on these extensive 3D datasets may result in a lack of appearance information for specific objects.
One potential solution is to further fine-tune our color diffusion model on real images, which we leave for future exploration.

Secondly, the current model demonstrates limitations in compositionality. 
Although \OURS can generate multiple objects with close relations, such as ``a house with a garden", it struggles to generate novel combinations like ``an astronaut riding a horse".
This issue is also observed in previous methods such as Shap-E~\cite{shape}.
We attribute this limitation to two main factors:
(1) The scarcity of multiple objects in a single CAD model contributes to the difficulty of generating diverse objects within one tri-plane. 
Unlike 2D images, where multiple objects are commonly present, most 3D CAD models consist of either a single object or two or three highly related objects.
(2) Current 3D datasets are still orders of magnitude smaller than their 2D counterparts, resulting in insufficient training data to effectively learn novel compositionality.

{
    \small
    \bibliographystyle{ieeenat_fullname}
    \bibliography{main}
}

\end{document}